\pdfoutput=1

\documentclass[11pt]{article}

\usepackage[]{emnlp2021}

\usepackage{times}
\usepackage{latexsym}

\usepackage{amssymb}
\usepackage{booktabs}
\usepackage{natbib}
\usepackage{enumitem}
\usepackage{hyperref}
\usepackage{subcaption}
\usepackage{graphicx}
\usepackage{multirow}
\usepackage{amsmath}

\usepackage[final]{changes}

\usepackage[T1]{fontenc}

\usepackage[utf8]{inputenc}

\usepackage{microtype}

\title{Breaking BERT: Understanding its Vulnerabilities for Named Entity Recognition through Adversarial Attack}

\author{Anne Dirkson \\\And
  Suzan Verberne \\
  Leiden Institute of Advanced Computer Science, Leiden University \\
  Niels Bohrweg 1, 2333 CA Leiden, the Netherlands \\
  \texttt{\{a.r.dirkson, s.verberne, w.kraaij\}@liacs.leidenuniv.nl} \\\And
  Wessel Kraaij \\
}

\begin{document}
\maketitle
\begin{abstract}
Both generic and domain-specific BERT models are widely used for natural language processing (NLP) tasks.
In this paper we investigate the vulnerability of BERT models to variation in input data for Named Entity Recognition (NER) through adversarial attack. Experimental results show that BERT models are vulnerable to variation in the entity context with 20.2 to 45.0\% of entities predicted completely wrong and another 29.3 to 53.3\% of entities predicted wrong partially. BERT models seem most vulnerable to changes in the local context of entities and often a single change is sufficient to fool the model. The domain-specific BERT model trained from scratch (SciBERT) is  more vulnerable than the original BERT model or the domain-specific model that retains the BERT vocabulary (BioBERT). We also find that BERT models are particularly vulnerable to emergent entities. Our results chart the vulnerabilities of BERT models for NER and emphasize the importance of further research into uncovering and reducing these weaknesses.
\end{abstract}



\section{Introduction}

Self-attentive neural models, such as BERT \citep{Devlin2018}, attain a high performance on a wide range of natural language processing (NLP) tasks. Despite their excellent performance, the robustness of BERT-based models is contested: Various studies \cite{Jin2019, Hsieh2019, Li2020, Zang2020, Sun2020} recently showed that BERT is vulnerable to adversarial attacks. Adversarial attacks are deliberate attempts to fool the model into giving the incorrect output by providing it with carefully crafted input samples, also called adversarial examples.

At present, the work on adversarial attack of Named Entity Recognition (NER) models is limited to a single study: \citet{araujo2020adversarial} attack biomedical BERT models by simulating spelling errors and replacing entities with their synonyms. They find that both attacks drastically reduce performance of these domain-specific BERT models on medical NER tasks.


Here, we aim to systematically test the robustness of BERT models for  NER under severe stress conditions in order to investigate \textit{which} variation in entities and entity contexts BERT models are most vulnerable to. This will, in turn, further our understanding of what these models do and do not learn. To do so, we propose two adversarial attack methods: replacing words in the context of entities with synonyms, and replacing entities themselves with others of the same type. In contrast to previous work, the methods we propose are adaptive and specifically target BERT's weaknesses: We create adversarial examples by making the changes to the input that either manage to fool the model or bring it closest to making a mistake (i.e. lower the prediction score for the correct output) instead of randomly introducing noise or variation.

We address the following research questions: 

\begin{enumerate}
	\item{How vulnerable are BERT models to adversarial attack on general and domain-specific NER?}
	\item{To what extent is the vulnerability impacted by domain-specific training?}
	\item{To which types of variation are BERT models for NER the most vulnerable?}
\end{enumerate}

Designing methods for direct adversarial attack of NER models poses additional challenges compared to the attack of text classification models as labels are predicted per word, sentences can contain multiple entities and entities can contain multiple words. 
To ensure that labels remain accurate in our adversarial sentences, we constrain synonym replacements to non-entity words when altering the context of the entity (i.e. an \emph{entity context attack}) and substitute entities only by entities of the same type when attacking the entity itself (i.e. an \emph{entity attack}). In line with previous work \citep{Jin2019, Li2020}, we include a minimal semantic similarity threshold based on the Universal Sentence Encoder \cite{Cer2018} to safeguard semantic consistency. Nonetheless, we acknowledge that for entity replacement adversarial examples may not be semantically consistent (e.g. if ``Japan'' is replaced with ``China'' in the sentence ``Tokyo is the capital of Japan''). Although factually incorrect, the resulting sentences can be considered utility-preserving i.e. they retain their usefulness as valid input to the model \citep{Sanchez2014}, because BERT models should be able to identify that the final word in the sentence is a country even if it is not the correct country. In real-world data, sentences are not necessarily factually correct.


We assume a black-box setting, which means that the adversarial method has no knowledge of the data, parameters or model architecture \citep{Alzantot2018}. This allows our methods to also be used for other neural architectures. Although we use English data, our methods are largely language-independent. Only an appropriate language model for synonym selection would be required.

The contributions of this paper are twofold: We adapt existing adversarial attack methods to sequence labelling tasks and evaluate the vulnerability of general and domain-specific BERT models for NER. We make our code available for follow up research.\footnote{Our code (BSD-3 Clause license), URLs to the benchmark data and the annotation guideline are available at: \url{https://github.com/AnneDirkson/breakingBERT}}

\section{Related Work}

In prior work, token-level black-box methods for adversarial attack have mainly been developed for classification and textual entailment \cite{Alshemali2019}. Substituting tokens with their synonyms 
is the most popular choice for perturbing at the token level. 
Synonyms are often found using nearest neighbours in a word embeddings model. One major challenge when selecting synonyms based on word embeddings is that antonyms will also be close in the embedding space. To solve this issue, recent studies \citep{Jin2019, Li2019, Hsieh2019} require a minimal semantic similarity between  the generated and original sentence. Additionally, some methods \citep{Alzantot2018, Jin2019} use word embeddings with additional synonymy constraints \cite{Mrksic}. We will employ both techniques. 




Approaches also differ in how they select the word that is perturbed: while some select words randomly, it is more common to use the importance of the word for the output \cite{Alshemali2019}. The importance is often operationalized as the difference in output before and after removing the word. We follow this approach in our method.




Most adversarial attack methods were developed for attacking recurrent neural models. However, there has been a growing interest in attacking self-attentive models in the last year 
\cite{Sun2020, Hsieh2019, Jin2019, Li2020, Zang2020, araujo2020adversarial, Balasubramanian2020, mondal-2021-bbaeg}. 
Nonetheless, the only study that has attacked BERT models for NER is the study by \citet{araujo2020adversarial}. They perform two types of character-level (i.e. swapping letters and replacing letters with adjacent keys on the keyboard) and one type of token-level perturbation (i.e. replacing entities with their synonyms). The authors find that biomedical BERT models perform far worse on NER tasks when spelling mistakes are included or synonyms of entities are used. 

Our work differs from \citet{araujo2020adversarial} in three ways. First, our adversarial examples are generated based on the importance of words for the correct output instead of through random changes. Thereby, we are able to test the robustness of BERT on the most severe stress conditions, while \citet{araujo2020adversarial} evaluate the scenario where the input data is noisy due to spelling mistakes and use of synonyms. Second, we analyze the impact of replacing entities with others of the same type (e.g. `France' with `Britain') and replacing words in the context of entities (see Table \ref{tab:aimex} for an example) instead of replacing entities with their synonyms. Third, we will test our method on the original BERT model as well as biomedical BERT models and on both generic and biomedical NER.

\begin{table*}[ht]
    \centering
    \small
    \begin{tabular}{cccccc}
        The & \underline{Republic} & \underline{of} & \underline{China} & \textbf{bought} & flowers\\
        $<$O$>$ & $<$B-LOC$>$ &  $<$I-LOC$>$ &  $<$I-LOC$>$ & $<$O$>$ &$<$O$>$ \\
        The & \underline{Republic} & \underline{of} & \underline{China } & \textbf{purchased} & flowers\\
        $<$O$>$ & $<$O$>$ &  $<$O$>$ &  $<$B-LOC$>$ & $<$O$>$ &$<$O$>$ \\
        
    \end{tabular}
    \caption{Example of a partial success. The \textbf{bold} word has been changed to attack the entity `Republic of China'.}
    \label{tab:aimex}
\end{table*}

\section{Methods} 
In this section, we describe two methods for generating adversarial examples designed to fool NER models, namely through (1) synonym replacements in the entity context (\emph{entity context attack}) and (2) entity replacement (\emph{entity attack}). These are described in Sections~\ref{sec:context} and~\ref{sec:entity}, respectively. 





\subsection{Aim of the attacks}
We aim to generate adversarial examples in which a target entity is no longer recognized correctly. This can be either because it has become a false negative or it has been assigned a different entity type. The attack is considered a success when the correct label has been changed, unless it has changed from the I-tag to the B-tag of the same entity type under the IOB schema. An example of this can be seen in Table \ref{tab:aimex}: Here, the start of the entity is mislabelled but the last part of the entity is still recognized. We consider this a partial success. 

\subsubsection{Metrics for evaluation of the attacks} 
The success of the attack and thus the vulnerability of the model is evaluated by the percentage of entities that were originally correctly labelled but are mislabelled after attack. For \textit{entity context attacks} entities can also be partially mislabelled i.e. only some words in the entity are mislabelled. This is captured by the partial success rate: the percentage of entities for which not the whole entity but at least half of the entity is mislabelled. For context attacks we also include a metric (`Words perturbed') to measure how much the sentence needed to be changed before the attack was successful: the average percentage of words that were perturbed out of the total amount of out-of-mention words in the sentences. This metric functions as a proxy for how difficult it is to fool the model \citep{Jin2019}.


\subsubsection{Entity Context Attack}\label{sec:context} 

To investigate the impact of the context on the correct labelling of the entity, we adapt the method of \citet{Jin2019}, which was designed for text classification, to sequence labelling tasks. For each entity in the sentence, a separate adversarial example is created, as models may rely on different contextual words for different entities. 

\paragraph{Step 1: Choosing the word to perturb} 
We use the importance ranking function shown in Equation \ref{eq:imps} to rank words based on their importance for assigning the correct label to the entity. The importance ($I_{w}$) of a word $w$ for a token in the entity is calculated as the change in the predictions (logits\footnote{Here logits refers to the vector of raw (non-normalized) predictions that the BERT model generates}) of the \textit{correct} label before and after deleting the word from the sentence \cite{Jin2019}. If the deletion of the word leads to an \textit{incorrect} label for the entity token, the importance of the word is increased by also adding the raw prediction score (logits) attributed to the incorrect label. 

If the entity consists of multiple words, we rank words based on their summed importance for correctly labelling each of the individual words in the entity. Besides stop words, we also exclude other entities from being perturbed. We adapt the function so that for any word with an I-tag, both the I and B label of the entity type (e.g. B-PER and I-PER) are considered correct.

Given a sentence of $n$ words $X = {w_1, w_2, . . . , w_n}$, the importance ($I_{w}$) of a word $w$ for a token in the entity is formally defined as: 


         
        
    

\begin{equation} \label{eq:imps}
\scriptsize
\begin{split}
I_{w}  = & F_{Y} (X) - F_{Y}(X_{-w}) \\
 & \textrm{if} F(X_{-w}) = Y \lor (F(X) = Y_{I} \land F(X_{-w}) = Y_{B})\\
 & F_{Y}(X) - F_{Y}(X_{-w})+ F\_{\bar{Y}}(X_{-w}) - F_{\overline{Y}}(X)\\
 & \textrm{if} F(X_{-w}) \neq Y \\
\end{split}
\end{equation}

where $F_{Y}$ is the prediction score for the correct label,  $F_{\bar{Y}}$ is the prediction score of the predicted label, $F$ is the predicted label, $Y$ is the correct label, $Y_{I}$ is the I-tag version of the correct label, $Y_{B}$ is the B-tag version of the correct label and $X_{-w}$ is the sentence X after deleting the word $w$.

\begin{table*}[t]
\centering
\small
\begin{tabular}{llp{5cm}cccp{2.2cm}}
\toprule
Domain & Dataset & Entity types & Dev. & Train & Test & Subset for Eval.* (\# of Entities)  \\ \midrule
 General & CoNLL-2003 & Person, Location, Organization, Miscellaneous  & 3,466 & 14,987& 3,684 &  500 (1,343)\\ \midrule
General & W-NUT 2017 & Person, Location,  Corporation, Product, Creative-work, Group  & 1,008 & 1,000 & 1,287 & 500 (787) \\  
 \midrule
Biomedical & BC5CDR & Disease, Chemical &  4,580 & 4,559  & 4,796 & 500 (1,221) \\ \midrule
Biomedical & NCBI disease & Disease &922 &5,432& 939 & 487 (897) \\ 
\bottomrule
\end{tabular}
\caption{Size of the data sets (number of sentences). *This subset is used for automatic evaluation}
\label{tab:size}
\end{table*}

\paragraph {Step 2: Gathering synonyms}

For each word, we select synonyms from the Paragram-SL999 word vectors \cite{Mrksic} with a similarity to the original word above the threshold $\delta$. \citet{Mrksic} injected antonymy and synonymy constraints into the vector space representation to specifically gear the embeddings space towards synonymy. These embeddings achieved state-of-the-art performance on SimLex-999 \citep{Hill2015} and were also used by \citet{Jin2019} and \citet{Alzantot2018}. We chose 0.5 as the minimal similarity threshold $\delta$ for synonym selection in contrast to the threshold of 0 used by previous work in order to better guarantee semantic similarity. Regardless of $\delta$, a maximum of 50 synonyms are selected. Examples of word pairs above a $\delta$ of 0.5 are `bought' and `obtained'; and `cat' and `puss'.
Below this threshold but within the first 50 synonyms fall `bought' and `forfeited'; and `cat' and `dustpan'.




\paragraph{Step 3: Filtering synonyms}
To preserve syntax, synonyms must have the same POS tag as the original word. If the data did not include POS tags, we added POS tags using NLTK. We filter the generated sentences for a sufficiently high semantic similarity to the original sentence. Semantic similarity is calculated with the Universal Sentence Encoder (USE) \citep{Cer2018}. 
We exclude synonyms that result in sentences falling below the similarity threshold $\epsilon$
. 

\paragraph{Step 4: Selecting the final synonym}

After filtering, we check whether any of the synonyms can change the entity label(s) fully. If there are multiple options, we select the one that leads to the highest sentence similarity ($\epsilon$) to the original sentence. If there are none, we select the synonym which can reduce the (summed) prediction scores of the correct label(s) the most. If no synonyms are left after filtering or none manage to reduce the prediction scores, we do not replace the original word. 

For multi-word entities, it is possible that a synonym changes some, but not all, labels. From the synonyms that change the most labels, we select the one that leads to the largest reduction in 
the (summed) prediction scores for the unchanged labels (i.e. the labels that are still predicted correctly  by the model) (see Equation \ref{eq:imps}). Which labels are still correct can differ per synonym. 

\paragraph{Finalizing the adversarial examples} 
For each word in this ranking, we go through step 2-4 until either the label(s) of the entity have been changed fully or there are no words left to perturb. Once the attack is partially successful, only the predictions of the not yet incorrectly labelled words in the entity are considered for subsequent iterations. 







\subsubsection{Entity Attack} \label{sec:entity}
To explore to what extent models rely on the words of the entity itself, we replace the entity with one of the same type e.g. we change `Japan' to another location. If a sentence contains multiple 
entities, an adversarial sentence is generated for each entity. The replacement entity is selected from a list of all entities in the data that are of the same type. We randomly select 50 candidate replacements from the entity list. We exclude candidates that result in a sentence that is too semantically dissimilar from the original (i.e. falling below the semantic similarity threshold $\epsilon$). For the remaining candidate entities, we check if the predicted label is incorrect. If so, we select the successful attack replacement with the highest semantic similarity at the sentence level. If not, the attack was unsuccessful.

\begin{table*}[t]
\centering
\small
\begin{tabular}{@{}lllllllll@{}}
\toprule
 & CoNLL-2003& W-NUT 2017 & NCBI-disease & BC5CDR \\
 \midrule
& \multicolumn{1}{c}{BERT} & \multicolumn{1}{c}{BERT}& \multicolumn{1}{c}{BERT} & \multicolumn{1}{c}{BERT}\\ \midrule
Success rate (\%) & 36.3 $\pm$ 0.612 & 42.2 $\pm$ 0.677 & 20.2 $\pm$ 0.443 & 38.8 $\pm$ 0.862 \\
Of which: & & & & \\
-- Missed entity (\%) & 47.4 $\pm$ 2.9 & 61.3 $\pm$ 5.1 & 100 & 90.4 $\pm$ 4.2  \\
-- Entity type error (\%) &  52.6 $\pm$ 2.9 & 38.7 $\pm$ 5.1 & 0 & 9.6 $\pm$ 4.2 \\

Partial success rate (\%) & 51.0 $\pm$ 0.465 & 51.6 $\pm$ 1.6 & 29.3 $\pm$ 0.841 & 45.9 $\pm$ 1.1 \\
Median semantic similarity & 0.928 $\pm$ 0.009 & 0.926 $\pm$ 0.017 & 0.920 $\pm$ 0.040 & 0.946 $\pm$ 0.002\\

Words perturbed (\%) & 15.6 $\pm$ 0.306 & 13.2 $\pm$ 1.2 & 12.4 $\pm$ 1.0 & 12.3 $\pm$ 0.04\\
\bottomrule
\end{tabular}
\caption{Automatic evaluation results for the context attacks on BERT models. Results are the mean of the three models}
\label{tab:attack}
\end{table*}

\begin{table*}[t]
\centering
\small
\begin{tabular}{@{}lllllllll@{}}
\toprule
 & \multicolumn{2}{c}{NCBI-disease} & \multicolumn{2}{c}{BC5CDR}\\               \midrule
& \multicolumn{1}{c}{BioBERT} & \multicolumn{1}{c}{SciBERT} &\multicolumn{1}{c}{BioBERT} & \multicolumn{1}{c}{SciBERT} \\ \midrule
Success rate (\%) & 20.9 $\pm$ 0.762 & 26.4 $\pm$ 0.875 & 37.9 $\pm$ 0.388 & 45.0 $\pm$ 0.665 \\
Of which: & & & & \\
-- Missed entity (\%) & 100 & 100 & 86.5 $\pm$ 2.3 & 87.1 $\pm$ 2.3 \\
-- Entity type error (\%) & 0 & 0 &  13.5 $\pm$ 2.3 & 12.9 $\pm$ 2.3\\

Partial success rate (\%) & 30.1 $\pm$ 1.032 & 39.0 $\pm$ 0.954 & 44.8 $\pm$ 0.331 & 53.3 $\pm$ 0.821 \\
Median semantic similarity & 0.921 $\pm$ 0.031 & 0.936 $\pm$ 0.030 & 0.921 $\pm$ 0.003 & 0.936 $\pm$ 0.008\\
Words perturbed (\%) & 9.1 $\pm$ 2.5 & 8.7 $\pm$ 2.9 & 9.8 $\pm$ 0.3 & 8.5 $\pm$ 0.8\\ \bottomrule
\end{tabular}
\caption{Automatic evaluation results for the context attacks on biomedical BERT models. Results are the mean of the three models}
\label{tab:attack2}
\end{table*}

\section{Experiments} 

\subsection{Data}

We use two general-domain English NER data sets for evaluating our method: the CoNLL-2003 data \citep{Tjong2003} and the W-NUT 2017 data \citep{Derczynski2017}. The goal of the latter was to investigate recognition of unusual, previously-unseen entities in the context of online discussions. Additionally, we use two English data sets from the biomedical domain: BC5CDR \citep{Li2016} and the NCBI disease corpus \citep{Dogan2014}. Both data sets have been used to evaluate domain-specific BERT models for NER in the biomedical domain \citep{Peng2019, Beltagy2019, Lee2019}. See Table \ref{tab:size} for more details on the data sets.



\subsection{Target models}

We fine-tune three BERT models (base-cased) for each data set with different initialization seeds (1, 2 \& 4) using the Huggingface implementation \citep{Wolf}. 
We set the learning rate at $5\times10^{-5}$ and optimized the number of epochs (3 or 4) as recommended in \citet{Devlin2018} for NER. We select the number of epochs based on the first BERT model (seed=1). We find that for all data sets except W-NUT 2017, 4 epochs is optimal. 


For the biomedical data sets, we additionally fine-tune two domain-specific BERT models BioBERT (base-cased) \cite{Lee2019} and SciBERT (scivocab-cased) \cite{Beltagy2019}. Each model is trained in three-fold (seeds are 1, 2 \& 4). 





\subsection{Evaluation of the adversarial attacks} 

\paragraph{Automatic evaluation\label{sec:auto}} 
We randomly select 500 eligible sentences from each test set. \added{Table \ref{tab:size} shows the number of entities in each subset.} We considered sentences to be eligible if they contain at least one entity and one verb. 
For the NCBI-disease data, only 487 sentences fulfill these criteria.

We use models trained on the original training and development data to perform NER on the selected subset of the test data. We then generate one adversarial example for each entity in the sentence that was initially predicted correctly. 
We evaluate to what degree models are fooled only for entities that were predicted correctly in the original sentences. We set the semantic similarity threshold at $\epsilon=0.8$ following \citet{Li2019}. Experiments are run on a GPU machine (NVIDIA Tesla K80). An experiment of three runs (one model on one data set) on one GPU will take roughly 20-24hrs. The models have 110 M parameters.




\paragraph{Human evaluation}

In order to evaluate the quality of our adversarial examples from the CoNLL-2003 and BC5CDR data, 100 original sentences and 100 adversarial sentences from each type of attack are scored for grammaticality by human judges. Grammaticality is evaluated on a five-point scale following the reading comprehension benchmark DUC2006 \citep{Dang}. Our annotators are four volunteering PhD students from our lab who have a background in linguistics\footnote{We opted for linguists as they are more acquainted with assessing grammaticality than biomedical domain experts}: two for each data set with 20\% overlap. We choose to present annotators with different original sentences than the ones on which the adversarial sentences they evaluate are based to prevent bias.




\section{Results}

    

\subsection{Entity Context Attack \label{sec:co}} 

\begin{figure*}[htb]
	\begin{subfigure}{.45\textwidth}
		\includegraphics[width=\textwidth]{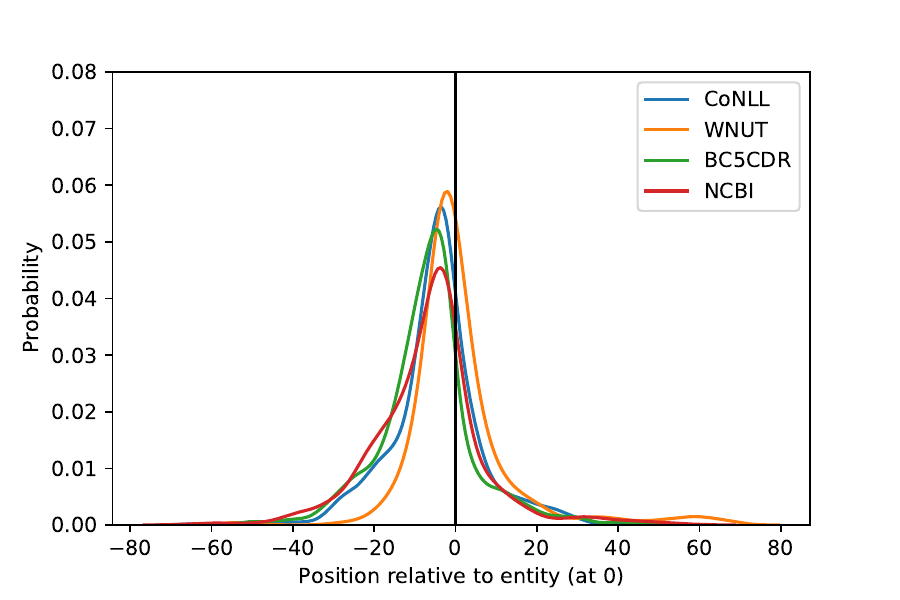}
		\caption{BERT}
		\label{fig:pertbert}
	\end{subfigure}
	\begin{subfigure}{.45\textwidth}
		\includegraphics[width=\textwidth]{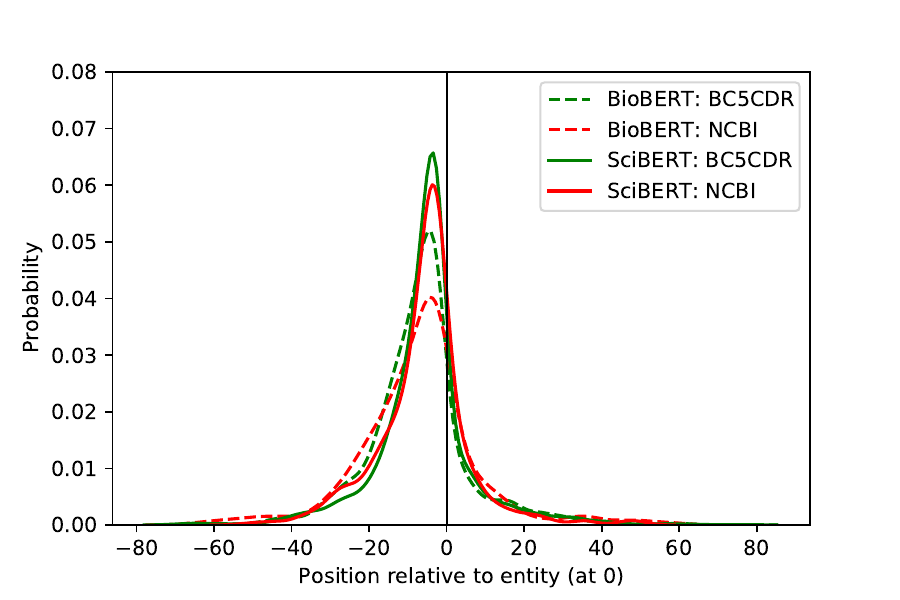}
		\caption{BioBERT and SciBERT}
		\label{fig:pertbio}
	\end{subfigure}
\caption{Distance of successful synonym replacements relative to the entity (at 0)}
\label{fig:pert}
\end{figure*} 





With an adversarial context attack, BERT models can be fooled into predicting entities partially or fully wrong (Partial + full success rate) for 87.3\% and 93.8\% of entities for CoNLL and W-NUT respectively. Moreover, for over 75\% of the cases the BERT models were fooled by a single change.
SciBERT appears more vulnerable than BERT, both to completely being fooled (+6.2 and +6.2\% point) and being fooled partially (+9.7 and +7.4 \% point) by context attack. Also the domain-specific models were often fooled by only one word being replaced with its synonym; BioBERT was fooled by a single change 65 and 75\% of the time whereas SciBERT was fooled by a single change 68 and 76\% of the time. 


We analyzed the sentence statistics for successful and failed attacks. Specifically for BERT models, we see that the following cases are more vulnerable to attacks: longer sentences; sentences with more words that could be replaced by synonyms; and shorter entities. Manual analysis of successful attacks reveals that BERT models are vulnerable when common words are replaced by rare synonyms (e.g. replacing `healthy' by `salubrious').

Figure \ref{fig:pertbert} shows where in the sentence changes have occurred in order to fool BERT. BERT models seem most vulnerable to changes in the local context of entities: only 1-2 words left or right of the entity. Manual analysis revealed that these words are often verbs. Although less influential, long distance context does appear to be used for predicting entities in some cases. We manually inspected sentences with long distance changes (\textgreater 20 words). Lists stood out as a prime example of a sentence type for which long distance context is important (e.g. ``The ministry said the group consisted of 13 nuns, seven Italians, and six Zaireans, and four priests, two from Belgium, one from Spain and one from Zambia.'').

For the BioBERT model, the distribution is strikingly similar to that of the original BERT model (see Figure \ref{fig:pertbio}). This is likely due to either the vocabulary or the training data\footnote{BioBERT includes all the original BERT training data as well as additional domain-specific data} that these models share. SciBERT models which share neither training data nor vocabulary with the original BERT model are even more vulnerable to changes in the local context of the entity (see Figure \ref{fig:pertbio}).

\subsection{Evaluating the necessity of importance ranking \label{sec:imp}} 
 
To investigate the effect of adding the word importance ranking to the entity context attack we perform an ablation study on the CoNLL-2003 test set. As can be seen in Table \ref{tab:random}, removing the word importance ranking leads to a stark drop in both the average full success of adversarial attack (from 37.3\% to 9.5\%) 
and the average partial success rate (from 52.8\% to 20.1\%). The number of words that need to be perturbed also drops, by 6.9\% point, meaning that attacks need less changes on average to be successful. Thus, it appears that the word importance ranking is crucial to the success of the adversarial attack algorithm.


\begin{table}[htb]
\centering

\centering
\small
\begin{tabular}{@{}lp{2.2cm}ll@{}}
\toprule
& \multicolumn{2}{c}{Importance ranking} \\
                             & Yes                                              & No                                               \\ \midrule
Success rate (\%)       & 37.3 $\pm$ 0.515 & 9.5 $\pm$ 4.3                                           \\
Partial success rate (\%) & 52.8 $\pm$ 0.356 & 20.1 $\pm$ 9.2  \\
Semantic similarity          & 0.922 $\pm$ 0.006  & 0.983 $\pm$ 0.006   \\
Words perturbed  (\%)   & 13.8 $\pm$ 0.238\                                           & 6.9 $\pm$ 2.2\                                             \\ \bottomrule
\end{tabular}
\caption{Comparison of context attacks  with and without importance ranking on CoNLL-2003 data}
\label{tab:random}
\end{table}



\subsection{Entity Attack \label{ent}}  

\begin{table*}[t]
\small
\centering
\begin{tabular}{@{}lllll@{}}
\toprule
&  CoNLL-2003& W-NUT 2017 & NCBI-disease & BC5CDR \\ \midrule 
& \multicolumn{1}{c}{BERT} & \multicolumn{1}{c}{BERT} & \multicolumn{1}{c}{BERT}& \multicolumn{1}{c}{BERT} \\ \midrule
Success rate (\%) & 97.5 $\pm$ 0.037 & 89.5 $\pm$ 0.886 & 99.2 $\pm$ 0.114 & 99.4 $\pm$ 0.073\\
Of which: & & \\
-- Missed entity(\%) & 21.3 $\pm$ 12.3 &  71.4 $\pm$ 1.9 & 100 & 86.1 $\pm$ 0.5  \\
-- Entity type error(\%) & 78.8 $\pm$ 12.3 & 28.6 $\pm$ 1.9 & 0 & 13.9 $\pm$ 0.5 \\
Median semantic similarity & 0.959 $\pm$ 0.001 &  0.928 $\pm$ 0.003 & 0.952 $\pm$ 0.001 & 0.962 $\pm$ 0.000\\ 
\bottomrule

\end{tabular}

\caption{Automatic evaluation results for entity attacks on BERT models. Results are the mean of three models.}
\label{tab:entattack}
\end{table*}

\begin{table*}[t]
\small
\centering
\begin{tabular}{@{}lllll@{}}
\toprule
&  \multicolumn{2}{c}{NCBI-disease} & \multicolumn{2}{c}{BC5CDR}\\ \midrule
& \multicolumn{1}{c}{BioBERT} & \multicolumn{1}{c}{SciBERT} & \multicolumn{1}{c}{BioBERT} & \multicolumn{1}{c}{SciBERT} \\ \midrule

Success rate (\%) & 99.2 $\pm$ 0.259 & 99.4 $\pm$ 0.054 & 99.4 $\pm$ 0.070 &  99.3 $\pm$ 0.089 \\
Of which: & & &&\\
-- Missed entity(\%) & 100 & 100 & 94.0 $\pm$ 0.7 & 91.7 $\pm$ 1.5 \\
-- Entity type error(\%) & 0 & 0 &  6.0 $\pm$ 0.7 & 8.3 $\pm$ 1.5 \\
Median semantic similarity & 0.955 $\pm$ 0.001 & 0.953 $\pm$ 0.002  & 0.961 $\pm$ 0.001 & 0.962 $\pm$ 0.001 \\
 \bottomrule
\end{tabular}
\caption{Automatic evaluation results for entity attacks on biomedical BERT models. Results are the mean of three models.}
\label{tab:entattack2}
\end{table*}




The main results of adversarial \emph{entity} attack on BERT models are presented in Table \ref{tab:entattack}. BERT models appear highly vulnerable to adversarial attacks on the entities themselves despite the high similarity between adversarial and original sentences. On average, BERT models are fooled for 97.5\% of entities that were initially predicted correctly on the CoNLL data and 89.2\% on W-NUT data. 
BERT models appear even more vulnerable to entity attacks on domain-specific data with success rates above 99\%.

Table \ref{tab:entattack2} shows that domain-specific BERT models 
do not resolve this issue. They are also highly vulnerable with over 99\% of all initially correctly predicted entities now predicted incorrectly. 
The high success rates of entity attacks both on general domain and domain-specific data suggests that BERT models, similar to traditional models, are unable to predict entities correctly based solely on the context of the entity. Replacing the entity word itself with another of the same entity type, with the context unchanged, can easily fool the model. This suggests a strong dependency on the entities that the model has seen previously, making these models vulnerable to new or emergent entities.  

This is corroborated by an analysis of which entities were chosen in successful attacks. For all BERT models and all datasets with the exception of the CoNLL data, these entities are significantly less frequent in training and development data than the original entities according to Wilcoxon signed rank tests  ($p<0.001$). 


A possible explanation for why BERT models for CoNLL are the exception is that there is a stronger match between the pretraining data and the data at hand than for the other data sets. This may make the model less vulnerable to infrequent entities, despite not being less vulnerable to entity replacement overall. 
Manual inspection further revealed that BERT models appear to be sensitive to the capitalization of entities (e.g. BERT models trained on CoNLL were fooled by transforming `New York' to `NEW YORK').

\subsection{Results of human evaluation \label{sec:human}}
On CoNLL-2003, the annotators have a fair inter-annotator agreement (weighted $\kappa=0.353$). On BC5CDR, the inter-annotator agreement is slight (weighted $\kappa=0.177$). Investigation of the annotations reveals that this is most likely because biomedical sentences are more difficult to assess for laymen. Because of the limited agreement we report grammaticality assessments per annotator. 

Table \ref{tab:gram} shows that although entity attacks do not significantly alter the grammaticality of the sentences, attacks on the context of the entity do. Although this reduction is consistent across data sets, the mean grammaticality of the adversarial sentences remains above 3 (acceptable) and the mean absolute reduction is less than a full point. 


\begin{table}[t]
\centering
\small
\begin{tabular}{@{}lllll@{}}
	\toprule
	& \multicolumn{2}{c}{CoNLL} & \multicolumn{2}{c}{BC5CDR} \\
	Annotator & \multicolumn{1}{c}{1} & \multicolumn{1}{c}{2} & \multicolumn{1}{c}{3} & \multicolumn{1}{c}{4}\\ 
	\midrule
	Original    &     3.51  & 4.34 & 4.43 & 4.78 \\
	After context attack & 3.05* & 3.68** & 3.86** & 4.35** \\
	After entity attack & 3.30 & 4.37 & 3.85 & 4.67    \\ \bottomrule
\end{tabular}
\caption{Mean grammaticality of the original and adversarial sentences. *$p<0.05$ **$p<0.01$ compared to the original sentences according to Mann-Whitney U tests. \vspace{-4mm}}
\label{tab:gram}
\end{table}

\section{Discussion and limitations}


We manually analysed the generated adversarial examples and found that our adversarial examples are susceptible to word sense ambiguity. For example, the top 50 synonyms for `surfed' in `surfed the Internet' includes both correct synonyms like `googled' and incorrect ones like `paddled'.
There are also some cases where the adversarial examples suffer from foreign words in the Paragram-SL999 word vectors \cite{Mrksic}. Occasionally synonyms are not English words (e.g. `number' to `nombre'), or synonym choice is influenced by words that occur in multiple languages e.g. `vie' in `to vie for top UN post' is replaced with `existence' which is a synonym of the French `vie' (i.e. life). 



Furthermore, our adversarial examples are susceptible to grammatical errors. Grammatically poor adversarial sentences often suffer from changes from verbs to nouns or vice versa that are not caught by the POS-filter (e.g. `open' to `openness' and `influence' to `implication'). These cases may be particularly difficult as `open' and `influence' can be both a verb and an adjective or noun. Another common error is singular-plural inconsistencies (e.g. `one dossiers'). To mitigate these issues, future work could focus on removing non-English words from the embedding space, and altering how the POS-tag of the synonym is determined. 



We find that semantic consistency can be an issue with broad entity types like location when attacking the entity itself. For example, in one case the country ``U.S.'' is replaced by the village ``Tavildara'' (in Tajikistan). For more specific entity types like Disease, Chemical or Person we do not encounter inconsistencies with sub-types of an entity category. On the contrary, often replacements are semantically close to the original. For instance, the anti-epileptic drug ``clonazepam'' was replaced by the anti-epileptic drug ``lorazepam'' and ``Washington'' in ``Washington administration'' was replaced by ``Clinton''.




Moreover, there are some caveats to keep in mind when interpreting weaknesses based on successful attacks. The architecture of self-attentive models means that the attention weight of a word is context-dependent. Thus, if changing that word fools the model, this might only be true in that context. Additionally, if multiple words were changed for a successful attack, their interactions may contribute to the success and it cannot simply be interpreted as caused by this combination of words. 





\section{Conclusions} 
We studied the vulnerability of BERT models in NER tasks under a black-box setting. Our experiments show that BERT models can be fooled by changes in single context words being replaced by their synonyms. 
They are even more vulnerable to entities being replaced by less frequent entities of the same type.

Our analysis of BERT's vulnerabilities can inform fruitful directions for future research. Firstly, our results reveal that rare or emergent entities remain a problem for both generic and domain-specific NER models. Consequently, we recommend further research into zero or few-shot learning. Moreover, the masking of entities during fine-tuning may be an interesting avenue for research. 
Secondly, BERT models also appear vulnerable to words it has not seen or rarely seen during training in the entity context. To combat this vulnerability, the use of adversarial examples designed specifically to include more infrequently used words could be explored. Another possible avenue for research could be alternative pre-training schemes for BERT such as curriculum learning \cite{elman1993learning}. 
Thirdly, we find that SciBERT is more vulnerable to changes in the entity context than BioBERT or BERT. This may be due to the domain-specific biomedical vocabulary that SciBERT employs, which could make it more vulnerable to out-of-entity words being replaced by more common English terms. This trade-off between robustness and domain-specificity of BERT models may be another worthwhile research direction. 

We consider our work to be a step towards understanding to what extent BERT models for NER are vulnerable to token-level changes and to which changes they are most vulnerable. We hope others will build on our work to further our insight into self-attentive models and to mitigate these vulnerabilities.

\bibliography{breakingbert}

\begin{thebibliography}{26}
\expandafter\ifx\csname natexlab\endcsname\relax\def\natexlab#1{#1}\fi

\bibitem[{Alshemali and Kalita(2019)}]{Alshemali2019}
Basemah Alshemali and Jugal Kalita. 2019.
\newblock \href {https://doi.org/10.1016/j.knosys.2019.105210} {{Improving the
  reliability of deep neural networks in NLP: A review}}.
\newblock \emph{Knowledge-Based Systems}, 10520.

\bibitem[{Alzantot et~al.(2018)Alzantot, Sharma, Elgohary, Ho, Srivastava, and
  Chang}]{Alzantot2018}
Moustafa Alzantot, Yash Sharma, Ahmed Elgohary, Bo-Jhang Ho, Mani~B Srivastava,
  and Kai-Wei Chang. 2018.
\newblock {Generating Natural Language Adversarial Examples}.
\newblock In \emph{Proceedings of the 2018 Conference on Empirical Methods in
  Natural Language Processing}, pages 2890--2896. Association for Computational
  Linguistics.

\bibitem[{Araujo et~al.(2020)Araujo, Carvallo, and
  Parra}]{araujo2020adversarial}
Vladimir Araujo, Andr{\'e}s Carvallo, and Denis Parra. 2020.
\newblock Adversarial evaluation of bert for biomedical named entity
  recognition.
\newblock In \emph{Proceedings of the The Fourth Widening Natural Language
  Processing Workshop}, pages 79--82.

\bibitem[{Balasubramanian et~al.(2020)Balasubramanian, Jain, Jindal, Awasthi,
  and Sarawagi}]{Balasubramanian2020}
Sriram Balasubramanian, Naman Jain, Gaurav Jindal, Abhijeet Awasthi, and Sunita
  Sarawagi. 2020.
\newblock \href {https://doi.org/10.18653/v1/2020.repl4nlp-1.24} {{What's in a
  Name? Are BERT Named Entity Representations just as Good for any other
  Name?}}
\newblock In \emph{Proceedings of the 5th Workshop on Representation Learning
  for NLP}, pages 205--214, Stroudsburg, PA, USA. Association for Computational
  Linguistics.

\bibitem[{Beltagy et~al.(2019)Beltagy, Lo, and Cohan}]{Beltagy2019}
Iz~Beltagy, Kyle Lo, and Arman Cohan. 2019.
\newblock \href {https://doi.org/10.18653/v1/D19-1371} {{S}ci{BERT}: A
  pretrained language model for scientific text}.
\newblock In \emph{Proceedings of the 2019 Conference on Empirical Methods in
  Natural Language Processing and the 9th International Joint Conference on
  Natural Language Processing (EMNLP-IJCNLP)}, pages 3615--3620, Hong Kong,
  China. Association for Computational Linguistics.

\bibitem[{Cer et~al.(2018)Cer, Yang, Kong, Hua, Limtiaco, St.~John, Constant,
  Guajardo-Cespedes, Yuan, Tar, Strope, and Kurzweil}]{Cer2018}
Daniel Cer, Yinfei Yang, Sheng-yi Kong, Nan Hua, Nicole Limtiaco, Rhomni
  St.~John, Noah Constant, Mario Guajardo-Cespedes, Steve Yuan, Chris Tar,
  Brian Strope, and Ray Kurzweil. 2018.
\newblock \href {https://doi.org/10.18653/v1/D18-2029} {Universal sentence
  encoder for {E}nglish}.
\newblock In \emph{Proceedings of the 2018 Conference on Empirical Methods in
  Natural Language Processing: System Demonstrations}, pages 169--174,
  Brussels, Belgium. Association for Computational Linguistics.

\bibitem[{Dang(2006)}]{Dang}
Hoa~Trang Dang. 2006.
\newblock {Overview of DUC 2006}.
\newblock In \emph{Proceedings of HLT-NAACL 2006}, pages 1--12.

\bibitem[{Derczynski et~al.(2017)Derczynski, Nichols, and {van
  Erp}}]{Derczynski2017}
Leon Derczynski, Eric Nichols, and Marieke {van Erp}. 2017.
\newblock {Results of the WNUT2017 Shared Task on Novel and Emerging Entity
  Recognition}.
\newblock In \emph{Proceedings ofthe 3rd Workshop on Noisy User-generated
  Text}, pages 140--147. Association for Computational Linguistics.

\bibitem[{Devlin et~al.(2019)Devlin, Chang, Lee, and Toutanova}]{Devlin2018}
Jacob Devlin, Ming-Wei Chang, Kenton Lee, and Kristina Toutanova. 2019.
\newblock \href {https://doi.org/10.18653/v1/N19-1423} {{BERT}: Pre-training of
  deep bidirectional transformers for language understanding}.
\newblock In \emph{Proceedings of the 2019 Conference of the North {A}merican
  Chapter of the Association for Computational Linguistics: Human Language
  Technologies, Volume 1 (Long and Short Papers)}, pages 4171--4186,
  Minneapolis, Minnesota. Association for Computational Linguistics.

\bibitem[{Doǧan et~al.(2014)Doǧan, Leaman, and Lu}]{Dogan2014}
Rezarta~Islamaj Doǧan, Robert Leaman, and Zhiyong Lu. 2014.
\newblock \href {https://doi.org/10.1016/j.jbi.2013.12.006} {{NCBI disease
  corpus: A resource for disease name recognition and concept normalization}}.
\newblock \emph{Journal of Biomedical Informatics}, 47:1--10.

\bibitem[{Elman(1993)}]{elman1993learning}
Jeffrey~L Elman. 1993.
\newblock Learning and development in neural networks: The importance of
  starting small.
\newblock \emph{Cognition}, 48(1):71--99.

\bibitem[{Hill et~al.(2015)Hill, Reichart, and Korhonen}]{Hill2015}
Felix Hill, Roi Reichart, and Anna Korhonen. 2015.
\newblock \href {https://doi.org/10.1162/COLI} {{SimLex-999: Evaluating
  Semantic Models With (Genuine) Similarity Estimation}}.
\newblock \emph{Computational Linguistics}, 41(4):665--695.

\bibitem[{Hsieh et~al.(2019)Hsieh, Cheng, Juan, Wei, Hsu, and
  Hsieh}]{Hsieh2019}
Yu-Lun Hsieh, Minhao Cheng, Da-Cheng Juan, Wei Wei, Wen-Lian Hsu, and Cho-Jui
  Hsieh. 2019.
\newblock {On the Robustness of Self-Attentive Models}.
\newblock In \emph{Proceedings of the 57th Annual Meeting ofthe Association for
  Computational Linguistics}, pages 1520--1529. Association for Computational
  Linguistics.

\bibitem[{Jin et~al.(2020)Jin, Jin, {Tianyi Zhou}, and Szolovits}]{Jin2019}
Di~Jin, Zhijing Jin, Joey~A {Tianyi Zhou}, and Peter Szolovits. 2020.
\newblock \href {http://arxiv.org/abs/1907.11932v2} {{Is BERT Really Robust? A
  Strong Baseline for Natural Language Attack on Text Classification and
  Entailment}}.
\newblock In \emph{AAAI}, page Accepted for publication.

\bibitem[{Lee et~al.(2019)Lee, Yoon, Kim, Kim, Kim, So, and Kang}]{Lee2019}
Jinhyuk Lee, Wonjin Yoon, Sungdong Kim, Donghyeon Kim, Sunkyu Kim, Chan~Ho So,
  and Jaewoo Kang. 2019.
\newblock \href {https://doi.org/10.1093/bioinformatics/btz682} {{BioBERT: a
  pre-trained biomedical language representation model for biomedical text
  mining}}.
\newblock \emph{Bioinformatics}.
\newblock Btz682.

\bibitem[{Li et~al.(2016)Li, Sun, Johnson, Sciaky, Wei, Leaman, Davis,
  Mattingly, Wiegers, and Lu}]{Li2016}
Jiao Li, Yueping Sun, Robin~J Johnson, Daniela Sciaky, Chih-Hsuan Wei, Robert
  Leaman, Allan~Peter Davis, Carolyn~J Mattingly, Thomas~C Wiegers, and Zhiyong
  Lu. 2016.
\newblock \href {https://doi.org/10.1093/database/baw068} {{BioCreative V CDR
  task corpus: a resource for chemical disease relation extraction}}.
\newblock \emph{Database}, 2016:68.

\bibitem[{Li et~al.(2019)Li, Ji, Du, Li, and Wang}]{Li2019}
Jinfeng Li, Shouling Ji, Tianyu Du, Bo~Li, and Ting Wang. 2019.
\newblock \href {https://doi.org/10.14722/ndss.2019.23138} {{TEXTBUGGER:
  Generating Adversarial Text Against Real-world Applications}}.
\newblock In \emph{Network and Distributed Systems Security (NDSS) Symposium
  2019}.

\bibitem[{Li et~al.(2020)Li, Ma, Guo, Xue, and Qiu}]{Li2020}
Linyang Li, Ruotian Ma, Qipeng Guo, Xiangyang Xue, and Xipeng Qiu. 2020.
\newblock \href {https://doi.org/10.18653/v1/2020.emnlp-main.500}
  {{Bert-attack: Adversarial attack against BERT using BERT}}.
\newblock In \emph{Proceedings of the 2020 Conference on Empirical Methods in
  Natural Language Processing}, page 6193–6202.

\bibitem[{Mondal(2021)}]{mondal-2021-bbaeg}
Ishani Mondal. 2021.
\newblock \href {https://doi.org/10.18653/v1/2021.naacl-main.423} {{BBAEG}:
  Towards {BERT}-based biomedical adversarial example generation for text
  classification}.
\newblock In \emph{Proceedings of the 2021 Conference of the North American
  Chapter of the Association for Computational Linguistics: Human Language
  Technologies}, pages 5378--5384, Online. Association for Computational
  Linguistics.

\bibitem[{Mrk\v{s}i\'{c} et~al.(2016)Mrk\v{s}i\'{c}, {\'{O} S\'{e}aghdha},
  Thomson, Ga\v{s}\'{i}c, Rojas-Barahona, Su, Vandyke, Wen, and Young}]{Mrksic}
Nikola Mrk\v{s}i\'{c}, Diarmuid {\'{O} S\'{e}aghdha}, Blaise Thomson, Milica
  Ga\v{s}\'{i}c, Lina Rojas-Barahona, Pei-Hao Su, David Vandyke, Tsung-Hsien
  Wen, and Steve Young. 2016.
\newblock {Counter-fitting Word Vectors to Linguistic Constraints}.
\newblock In \emph{Proceedings of NAACL-HLT 2016}, pages 142--148.

\bibitem[{Peng et~al.(2019)Peng, Yan, and Lu}]{Peng2019}
Yifan Peng, Shankai Yan, and Zhiyong Lu. 2019.
\newblock Transfer learning in biomedical natural language processing: An
  evaluation of bert and elmo on ten benchmarking datasets.
\newblock In \emph{Proceedings of the 2019 Workshop on Biomedical Natural
  Language Processing (BioNLP 2019)}.

\bibitem[{Sun et~al.(2020)Sun, Hashimoto, Yin, Asai, Li, Yu, and
  Xiong}]{Sun2020}
Lichao Sun, Kazuma Hashimoto, Wenpeng Yin, Akari Asai, Jia Li, Philip Yu, and
  Caiming Xiong. 2020.
\newblock \href {http://arxiv.org/abs/2003.04985} {{Adv-BERT: BERT is not
  robust on misspellings! Generating nature adversarial samples on BERT}}.
\newblock \emph{arXiv}.

\bibitem[{Sánchez et~al.(2014)Sánchez, Batet, and Viejo}]{Sanchez2014}
David Sánchez, Montserrat Batet, and Alexandre Viejo. 2014.
\newblock \href {https://doi.org/https://doi.org/10.1016/j.jbi.2014.06.008}
  {Utility-preserving privacy protection of textual healthcare documents}.
\newblock \emph{Journal of Biomedical Informatics}, 52:189 -- 198.

\bibitem[{Tjong Kim~Sang and {de Meulder}(2003)}]{Tjong2003}
Erik Tjong Kim~Sang and Fien {de Meulder}. 2003.
\newblock {Introduction to the CoNLL-2003 Shared Task: Language-Independent
  Named Entity Recognition}.
\newblock In \emph{Proceedings of the Seventh Conference on Natural Language
  Learning at HLT-NAACL 2003}, pages 142--147.

\bibitem[{Wolf et~al.(2019)Wolf, Debut, Sanh, Chaumond, Delangue, Moi, Cistac,
  Rault, Louf, Funtowicz, and Brew}]{Wolf}
Thomas Wolf, Lysandre Debut, Victor Sanh, Julien Chaumond, Clement Delangue,
  Anthony Moi, Pierric Cistac, Tim Rault, R{\'{e}}mi Louf, Morgan Funtowicz,
  and Jamie Brew. 2019.
\newblock \href {http://arxiv.org/abs/1910.03771v3} {{Transformers:
  State-of-the-art Natural Language Processing}}.
\newblock \emph{ArXiv}.

\bibitem[{Zang et~al.(2020)Zang, Hou, Qi, Liu, Meng, and Sun}]{Zang2020}
Yuan Zang, Bairu Hou, Fanchao Qi, Zhiyuan Liu, Xiaojun Meng, and Maosong Sun.
  2020.
\newblock \href {http://arxiv.org/abs/2009.09192} {{Learning to attack: Towards
  textual adversarial attacking in real-world situations}}.
\newblock \emph{arXiv}.

\end{thebibliography}
\bibliographystyle{acl_natbib}




\end{document}